\documentclass[conference]{IEEEtran}

\usepackage{cite}
\usepackage[table]{xcolor}
\usepackage{hhline}

\ifCLASSINFOpdf
  \usepackage[pdftex]{graphicx}
  \usepackage{epstopdf}
  \DeclareGraphicsExtensions{.pdf,.jpeg,.png,.eps}
\else
\fi

\usepackage[cmex10]{amsmath}
\usepackage{array}
\usepackage{tabularx}
\usepackage{multirow}
\usepackage[caption=false,font=footnotesize]{subfig}
\usepackage{url}

\begin{document}
%
\title{Vehicle Color Recognition using Convolutional Neural Network}


%
\author{\IEEEauthorblockN{Reza Fuad Rachmadi\IEEEauthorrefmark{1} and 
I Ketut Eddy Purnama\IEEEauthorrefmark{2}}
\IEEEauthorblockA{Department of Multimedia and Networking Engineering\\
Institut Teknologi Sepuluh Nopember,
Surabaya, Indonesia 60111\\ Email: \IEEEauthorrefmark{1}fuad@its.ac.id, \IEEEauthorrefmark{2}ketut@te.its.ac.id}\\} 


\maketitle

\begin{abstract}
Vehicle color information is one of the important elements in ITS (Intelligent Traffic System). In this paper, we present a vehicle color recognition method using convolutional neural network (CNN). Naturally, CNN is designed to learn classification method based on shape information, but we proved that CNN can also learn classification based on color distribution. In our method, we convert the input image to two different color spaces, HSV and CIE Lab, and run it to some CNN architecture. The training process follow procedure introduce by Krizhevsky, that learning rate is decreasing by factor of 10 after some iterations. To test our method, we use publicly vehicle color recognition dataset provided by Chen. The results, our model outperform the original system provide by Chen with 2\% higher overall accuracy.
\end{abstract}


%
\IEEEpeerreviewmaketitle

\section{Introduction}

Intelligent Transport System (ITS) is a system that manages transportation from traffic management to law enforcement. One important object that widely explored by ITS is a vehicle and their properties, including type, color, and license plate. Vehicle color is an important property for vehicle identification and provide visual cues for fast action law enforcement. Recognize vehicle color is very challenging task because several factors including weather condition, quality of video/image acquisition, and strip combination of the vehicle. The first factor, weather condition, may dramatically change the color illumination of the acquisition image. For example, if the image/video taken at haze condition then there a lot of "soft" white noise added to the image. Soft white noise means that the noise is not random but continues and blended with the foreground and background objects. The quality of video/image acquisition is affected the final decision of the vehicle color recognition system and its depends of the optical sensor in the camera. Camera that can capture object at high speed is recommended for ITS, but not all installed camera in the road can do that. A lot of cameras installed in the road only used to monitor the traffic, pedestrians, and street conditions. The last factor is strip combination of the vehicle, which is very affected to the vehicle recognition system. Region selection is very important to tackle the problem.

There are some research paper published to tackle vehicle color recognition problem, like in \cite{hsieh,pchen,dule,baek,jwson}. Chen et al. \cite{pchen} use feature context and linear SVM classifier to tackle the problem. Feature context is a collection of histogram that build with several areas, like spatial pyramid structure but with different region configuration. In other paper \cite{dule}, they try to tackle vehicle color recognition problem using 2D histogram with some ROI configuration as features and neural network as classifier. Baek et al. \cite{baek} also use 2D histogram but without ROI configuration and SVM as classifier. Another approach is described by Son et al. \cite{jwson} which using convolution kernel to extract similarity between positive and negative images and then feed up those similarity score to SVM classifier.

Color spaces are very important to color recognition applications, like vehicle color recognition. The selection of color space will impact the recognition performance. The most usable color space in digital photography is RGB color space, but RGB color space has problem to color recognition because channel of RGB color space contribute equal for each channel so to distinct color is more difficult. Usually, researcher will not use RGB as their primary color space and convert it to other color spaces that separate illumination and color, like CIE Lab or HSV \cite{baek,dule,jwson}. Another approach is to make 2D histogram of two channels, like H and S channel in HSV color space, and do classification using those 2D histogram.

In this paper, we present vehicle color recognition method using convolutional neural network (CNN). CNN is type of neural network but instead of using fully connected layer, CNN use layer called convolution layer to extract features from data. The training mechanism is very similar to normal neural network and use stochastic gradient descent as training algorithm. CNN is become very popular after winning the ILSVRC (ImageNet Large Scale Visual Recognition Challenge) 2012 \cite{alex}. In those paper, they use more than 600,000 neuron and 7 hidden layer to provide good model of the data. To avoid overfitting Krizhevsky et al. \cite{alex} employed regularization method called dropout to the fully connected layer \cite{srivastava}. The Krizhevsky model is huge and as reported in the paper, the model trained in six day for 450,000 iteration in GPU hardware. Before going into details, in section two we describe detils related works in color recognition. Section two describe details architecture of our CNN model. Section three reports the experiments we have done and discuss the results.

\begin{figure*}
\centering
\frame{\includegraphics[trim=-20 -10 -20 -10,width=1.\textwidth]{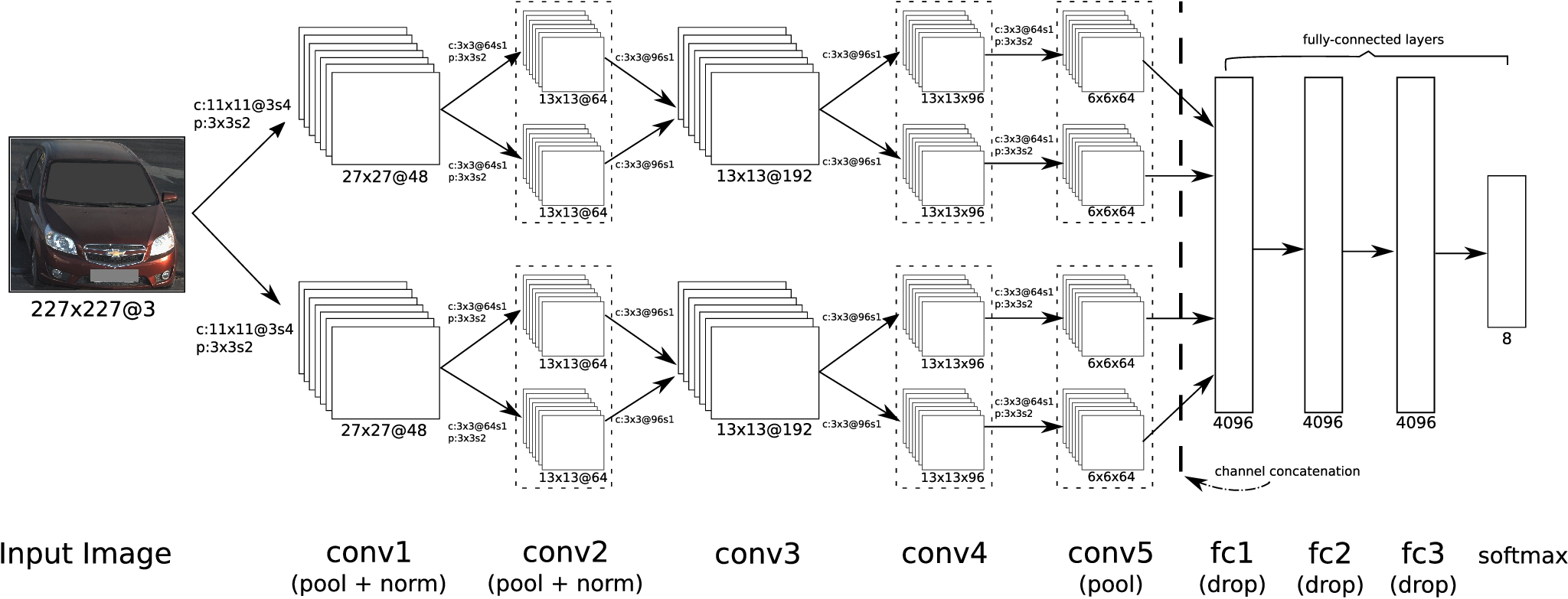}}
\caption{The CNN architecture used in our system consist 8 layers with 2 base networks with total 16 layers. First two layers and fifth layer does normalization and pooling after convolution process. The third and fourth layer does only convolution process. Before feed up to fully-connected layers, the networks do channel concatenation process. Sample of input image is taken from Chen \cite{pchen} dataset.}
\label{fig:cnnarch}
\end{figure*}

\section{Related Works}

There are several research that try to tackle vehicle color recognition problem including in \cite{hsieh,pchen,dule,baek,jwson}. The newest research is describe by Chen et al. \cite{pchen} in 2014 and Hsieh et al. \cite{hsieh} in 2015. Chen et al. use  feature context (FC) with selected configuration to divide the images into subregions, create histogram for each subregion, and learned it using linear SVM. Not all value in histogram is used to classify the vehicle color but the values clustered to form codebook for the problem and then choose the codebook as feature for the classifier. This mechanism know as Bag-of-Word (BoW) method. Chen et al. done preprocessing using haze removal method \cite{khe} and color contrast normalization method. The accuracy of system proposed by Chen et al. is very high, over 92\%. 

Another paper by Hsieh et al. \cite{hsieh} proposed color correction using background image and two frame image of car. Not only color correction method, Hsieh et al. also proposed window removal method that remove the window part of the car images and classify vehicle color using lower part, like bumper and doors, of the car. The window removal done by taking the orientation of the car, fit the detail segmented car image by ellipse shape and cut a half of the ellipse. Hsieh et al. done the experiments using three different classifier, G-Classifier, DC-Classifier, and DG-Classifier. G-Classifier responsible for classify gray and non-gray color. The method is very simple threshold method with assumption that for gray color the avarage of three channel, RGB, is very close with color value of each channel. The DC-Classifier and DG-Classifier trained using SVM with features extracted from RGB and CIE Lab color space. Red, green, blue, and yellow color class classified using DC-Classifier and the rest of the color class classified using DG-Classifier. From the experiments, Hsieh et al. report that the average accuracy for the system is 93,59\% with 7 color class including black, silver, white, yellow, red, green, and blue.

FC also used by Dule et al. \cite{dule} to tackle vehicle color recognition problem. The different between FC used by Chen et al. and Dule et al. is that Dule et al. only used two ROI (smooth hood peace and semi front vehicle). ROI is selected automatically using plate detection method and otsu thresholding to search smooth hood peace and heuristic approach for semi front vehicle. The classifier used by Dule et al. are K-NN, ANN, and SVM. The best accuracy that reported in Dule et al. paper is 83,5\% with configuration of 8 bin histogram, several combination of color spaces, and ANN classifier.   

Other approach for vehicle color recognition problem is classify vehicle color using 2D histogram features. Baek et al. \cite{baek} proposed the vehicle color recognition system using 2D histogram features and SVM classifier. Hue and saturation in HSV color space is used for creating the 2D histogram. From the experiments, the average accuracy of the system is 94,92\%. The dataset used in the experiment has 500 outdoor vehicle images with five color class including black, white, red, yellow, and blue color class.

Son et al. \cite{jwson} proposed other possible approach for color recognition using similirity method. The system using grid kernel that run on hue and saturation channel of HSV color space. The same dataset as in \cite{baek} is used in the experiments. Son et al. reported only precission and recall for each color class. The percentage of precission and recall from the experiments is very high and close to 100\%. High precission and high recall indicate that the model has good accuracy.

\section{The CNN Architecture}

The architecture of our CNN can viewed in figure \ref{fig:cnnarch}. Our CNN architecture consists 2 base networks and 8 layers for each base network with total 16 layers. The first two layers of our CNN architecture is a convlutional layer and it does convolution process following by normalization and pooling. Convolutional layer is a layer that do convolution process that same as convolution process in image processing algorithm. For $I_i$ is an input image and $h$ is a some convolution kernel, output image for convolution process $I_o$ can be written as
\begin{align}
	I_o[m,n] = \sum_{j=-\infty}^\infty\sum_{i=-\infty}^\infty I_i[i,j].h[m,n]
\end{align}
with $[m,n]$ is pixel value at coordinate $(m,n)$. Training process of CNN will learn $h$, may called as kernel, as parameters of convolutional layer. The choice of activation function in convolutional layer have huge impact for the networks. There a several choice of activation function including $tanh$ and $ReLU$ (Rectified Linear Unit). In our CNN networks we use $ReLU$ activation function for all layers including the fully-connected layers. The normalization process done by following equation \ref{eq:lrn} with $\alpha=10^{-4}$, $\beta=0.75$, and $n=5$.
\begin{align}
	l^i_{x,y} = k^i_{x,y} / \bigg(1 + \frac{\alpha}{n}\sum_{j=i-n/2}^{i+n/2} (k^i_{x,y})^2\bigg)^\beta
	\label{eq:lrn}	
\end{align}
with $l^i_{x,y}$ is normalization result and $k^i_{x,y}$ is output of layer activation function for convolution at coordinate $(x,y)$. Using those normalization, the accuracy of CNN increase about 2\% according to \cite{alex}. The last process in two first layers is pooling process. There are two type of pooling, max pooling and mean pooling. Each type has different approach, max pooling will take maximum respon from the convolutional process which is shape with sharp edges and mean pooling will take the average of the convolutional process respon which is summarize the shape in neighborhood. In our CNN architecture, we use max pooling with size 3x3 and stride 2 for overlapping pooling. The second, fourth and fifth layer are grouping into two group which each group is independent each others. The third and fourth layer is also a convolutional layer but without pooling and normalization process. Output of third and fourth layer is same as input because we use 3x3 kernel with pad 1 added for each border. The fifth layer is convolutional layer with only pooling process without normalization. 

Before going into a fully-connected layers, the pooling output of the fifth layer from two base networks is concatenate and flattened into one long vector. The sixth and seventh layer is a fully-connected layer employed dropout regularization method to reduce overfitting. The last layer is the softmax regression layer which can describe in the following equation
\begin{align}
	p(y^{(i)}=j|x^{(i)};\theta)=\frac{e^{\theta^T_jx^{(i)}}}{\sum_{l=1}^ke^{\theta^T_lx^{(i)}}}	
\end{align}
with $p(y^{(i)}=j|x^{(i)};\theta)$ is probability of $y^{(i)}$ being class $j$ given input $x^{(i)}$ with weight parameter $\theta$. 

Overall, our CNN architecture consists 2 base networks, 8 layers each with total 16 layers. First layer use 11x11@3 kernel with total 48 kernels, second layer use 3x3@48 kernel with total 128 kernels, third use 3x3@128 kernel with total 192 kernels, fourth layer use 3x3@192 kernel with total 192 kernels, and fifth layer use 3x3@192 with total 128 kernels. Pooling process is employed in first, second, and fifth layer with same parameter, pooling size of 3x3 with 2 pixel stride. Sixth, seventh, and eight layers is fully-connected layers with each 4096-4096-8 neuron with dropout regularization method employed in sixth and seventh layer. The network's input is a 3 channel image with 150,228 dimensional or 227x227@3 resolution. Total neuron involved in the networks is 658,280 neurons. 

\section{The Experiments}

\begin{figure}
\centering
\frame{\includegraphics[trim=-5 -10 -10 -10,width=.48\textwidth]{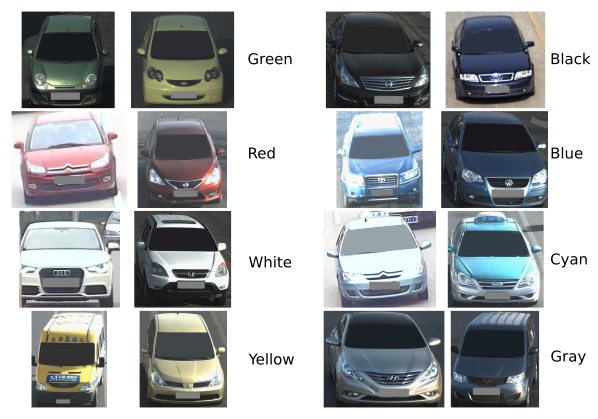}}
\caption{Sample images from Chen dataset \cite{pchen}. Some images are suffering from noise and brightness constancy.}
\label{fig:sample_dataset}
\end{figure}

\subsection{Training Process}

Our models trained using stochastic gradient descent with 115 examples per batch, momentum of 0.9 and weight decay of 0.0005. For the experiments, we use Chen dataset \cite{pchen} and some sample images of the dataset can be viewed in figure \ref{fig:sample_dataset}. The dataset contains 15601 vehicle images with 8 classes of vehicle color, which are black, blue, cyan, gray, green, red, white, and yellow. In the training process, half of class examples are used. Each example is resized into 256x256@3 resolution with certain color spaces. We use four different color spaces, RGB, CIE Lab, CIE XYZ, and HSV. Before the data processed for training, it cropped to 227x227@3 and subtracted by mean image of the training data. In training process the data randomly mirrored to increase the classifier accuracy. We use learning rate of 0.01 and reduced continuously by a factor of 10 at multiple iteration of 50,000 with maximum iteration of 200,000. We use caffe framework \cite{caffe} to implement our models. The weights of the networks are initialized using a gaussian function with $\delta=0.01$ for connecting weights and fixed value of $0.1$ for bias value. 

The stochastic gradient descent method, SGD for short, is an optimization method that want to find minimum or maximum value of some function. SGD will work for all function that have gradient or first derivative. Usually the system use SGD for minimizing the error or loss function and update the weight parameters based on following function
\begin{align}
	w_{i+1} = w_i - \alpha\nabla L(z,w_i)
\end{align} 
with $w_i$ is current weight parameters, $\alpha$ is learning rate, and $\nabla L(z,w_i)$ is the gradient of loss function $L$ with respect to input examples $z$. For faster model convergence, the weight decay and momentum are added to the update equation. The final equation of update function in SGD method is describe following
\begin{align}
	\Delta w_{i+1} &= \gamma \Delta w_i + (1-\gamma)(-\alpha\nabla L(z,w_i) \\
	w_{i+1} &= w_i - \alpha\nabla L(z,w_i) - \alpha\zeta w_i
\end{align}
with $\gamma$ is momentum variable and $\zeta$ is weight decay. Changing momentum and weight decay may accelerate the training process.

The training process done in GPU hardware to reduce the training time. Our GPU hardware consists 14 multiprocessor within 3 GB memory. There are two limitations of our GPU hardware for training process, the memory limiting the size of the networks or the batch size used in the training process and the maximum dimension of the grid block execution (parallel execution configuration) also limiting the batch size used in the training process. The training process taken over 2 GB GPU memory for the data and the networks with 4 days of execution time.

\begin{table}\footnotesize
\renewcommand{\arraystretch}{1.775}
\caption{Accuracy of our models with 4 different color spaces and accuracy from Chen et al. \cite{pchen} for comparation.}
\label{tbl:tbl_acc}
\centering
\begin{tabular}{c|>{\centering\arraybackslash}p{0.09\linewidth}|>{\centering\arraybackslash}p{0.09\linewidth}|>{\centering\arraybackslash}p{0.11\linewidth}|>{\centering\arraybackslash}c|>{\centering\arraybackslash}p{0.11\linewidth}|}
	\hhline{~-----|}
	& \multicolumn{4}{c|}{\cellcolor[gray]{0.85}\textbf{Color Space}} & \cellcolor[gray]{0.85}\\
	\hhline{----->{\arrayrulecolor[gray]{0.85}}->{\arrayrulecolor[rgb]{0,0,0}}}
	\multicolumn{1}{|c|}{\cellcolor[gray]{0.85}\textbf{Color Class}} & \multicolumn{1}{c|}{\cellcolor[gray]{0.85}RGB} & \multicolumn{1}{c|}{\cellcolor[gray]{0.85}HSV} & \multicolumn{1}{c|}{\cellcolor[gray]{0.85}CIE Lab} & \multicolumn{1}{c|}{\cellcolor[gray]{0.85}CIE XYZ} & \multirow{-2}{*}{\cellcolor[gray]{0.85}\parbox{1.\linewidth}{\centering Chen et al.\cite{pchen}}}\\
	\hline
	\multicolumn{1}{|r|}{yellow} & 0.9794 & 0.9450 & 0.9656 & \textbf{0.9828} & 0.9553\\
	\hline
	\multicolumn{1}{|r|}{white} & \textbf{0.9666} & 0.9624 & 0.9561 & 0.9649 & 0.9423\\
	\hline
	\multicolumn{1}{|r|}{blue} & 0.9410 & \textbf{0.9576} & 0.9410 & 0.9484 & 0.9535\\
	\hline
	\multicolumn{1}{|r|}{cyan} & 0.9645 & 0.9716 & 0.9645 & 0.9716 & \textbf{0.9787}\\
	\hline
	\multicolumn{1}{|r|}{red} & \textbf{0.9897} & 0.9866 & \textbf{0.9897} & 0.9886 & 0.9878\\
	\hline
	\multicolumn{1}{|r|}{gray} & 0.8608 & 0.8503 & \textbf{0.8668} & 0.8647 & 0.8466\\
	\hline
	\multicolumn{1}{|r|}{black} & \textbf{0.9738} & 0.9703 & 0.9703 & 0.9709 & 0.9730 \\
	\hline
	\multicolumn{1}{|r|}{green} & \textbf{0.8257} & 0.8215 & 0.8215 & 0.7676 & 0.7884 \\
	\hline
	\multicolumn{1}{|r|}{\cellcolor[gray]{0.85}\textbf{average}} & \cellcolor[gray]{0.85}\textbf{0.9447} & \cellcolor[gray]{0.85}0.9372 & \cellcolor[gray]{0.85}0.9414 & \cellcolor[gray]{0.85}0.9432 & \cellcolor[gray]{0.85}0.9282\\
	\hline
\end{tabular}
\end{table}

\begin{table}\footnotesize
\renewcommand{\arraystretch}{1.775}
\caption{Execution time for our model using CPU and GPU hardware.}
\label{tbl:tbl_exec}
\begin{tabular}{l|>{\centering\arraybackslash}p{0.15\linewidth}|>{\centering\arraybackslash}p{0.15\linewidth}|}
	\hhline{~--}
	& \cellcolor[gray]{0.85}CPU\newline(1 core) & \cellcolor[gray]{0.85}GPU\newline(448 cores) \\
	\hline
	\multicolumn{1}{|l|}{\cellcolor[gray]{0.85}\textbf{Initialization time}} & 4.849 s & 4.849 s \\
	\hline
	\multicolumn{1}{|l|}{\cellcolor[gray]{0.85}\textbf{Execution time}} & 3.248 s & 0.156 s \\
	\hline
\end{tabular}
\centering
\end{table}

\subsection{Results and Discussion}

\begin{figure}
	\centering	
	\includegraphics[trim=-50px 0px -50px 0px,width=1.\linewidth]{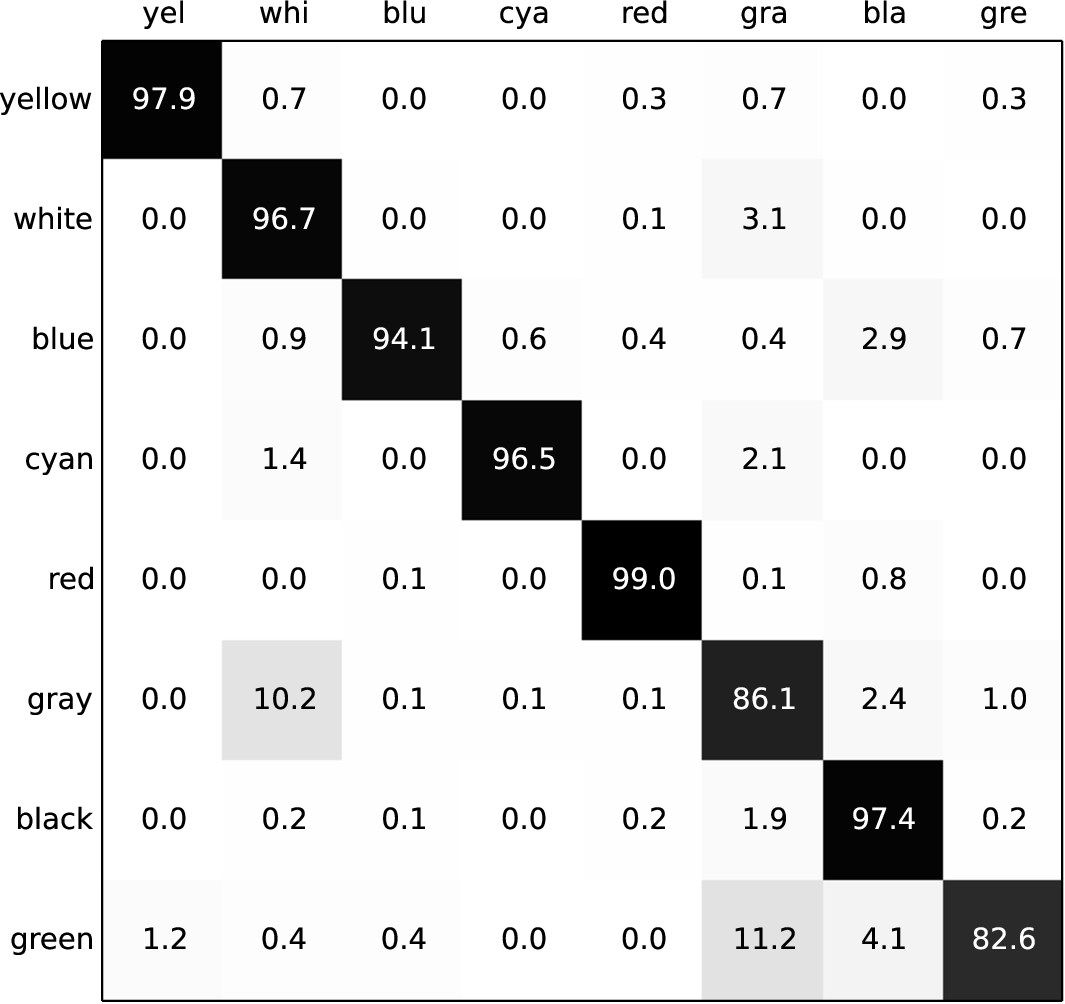}
	\caption{Confusion matrix from our model with RGB color space. Each cell describe accuracy for each class measure in percentage. }
	\label{fig:confmat}
\end{figure}

For testing purpose, we use 50\% examples of dataset that not used in the training process. Table \ref{tbl:tbl_acc} summarize our testing results with four different color spaces and compare the results with the system provide by Chen et al. \cite{pchen}. Each class consists different number of examples from 141 to 2371. From table \ref{tbl:tbl_acc}, it can see that RGB color space achieve the highest accuracy of the testing process with average accuracy of 94,47\%. Four color spaces used in the models have high accuracy, more than 90\%, with narrow deviation. The results show that our CNN model outperform the original system of dataset provide by Chen et al. \cite{pchen}. Our models outperform Chen et al. system in yellow, white, blue, red, gray, black, and green color class. Only on cyan color class our system had lower accuracy comparing to the Chen et al. system but with different only 0.7\%. Figure \ref{fig:confmat} is a confusion matrix for our model using RGB color space. The confusion matrix shows that the most worst accuracy of our model is in green and gray color class. Some examples of green class is misclassified as gray class and its above 10\%. As seen in the dataset, some green color class examples has color that very close to gray, more like green-gray color than green, so the classifier may have it wrong classified as a gray color class. The same case appears in gray color class which some gray color class examples misclassified as white color class. Thus may appears because of very bright sunlight reflection on metallic paint or the color is too light so it's very close to another color as well.

\begin{figure*}
	\centerline{
	\subfloat[Kernel from network 1]{
		\includegraphics[trim=-13px -33px -13px 0px,width=.225\linewidth]{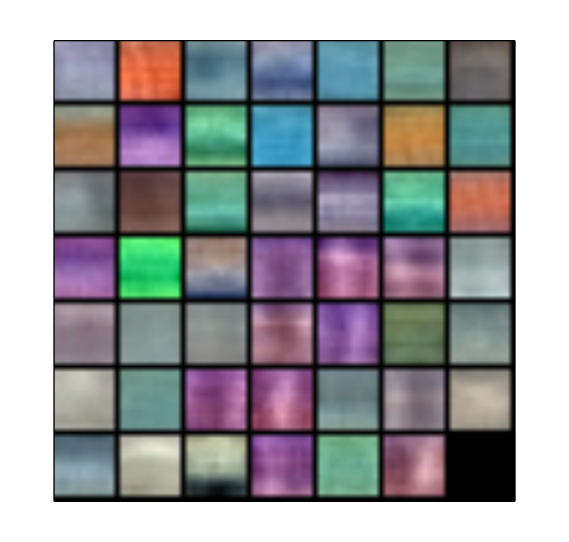}
		\label{fig:n1_conv1}}
	\hfil
	\subfloat[Kernel from network 2]{
		\includegraphics[trim=-13px -33px -13px 0px,width=.3\linewidth]{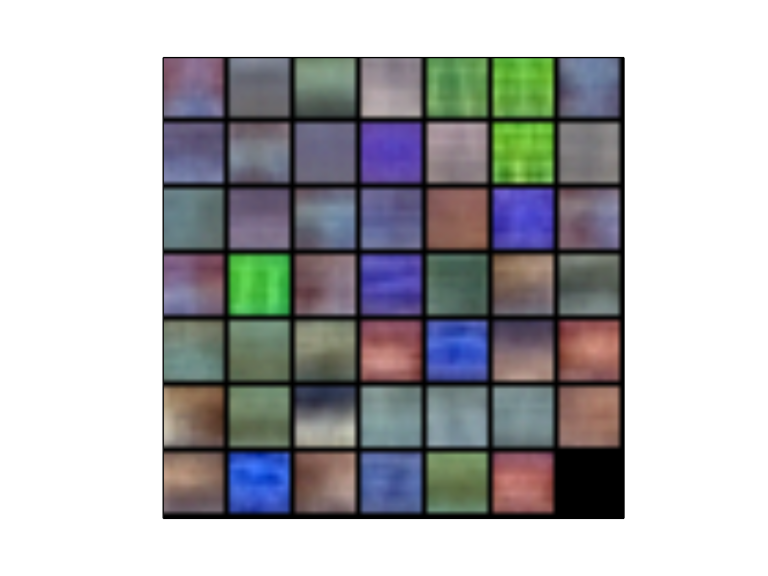}
		\label{fig:n1_conv6}}
	\hfil
	\subfloat[Output example from pooling process]{
		\includegraphics[trim=-10px 0px -10px 0px,width=.5\linewidth]{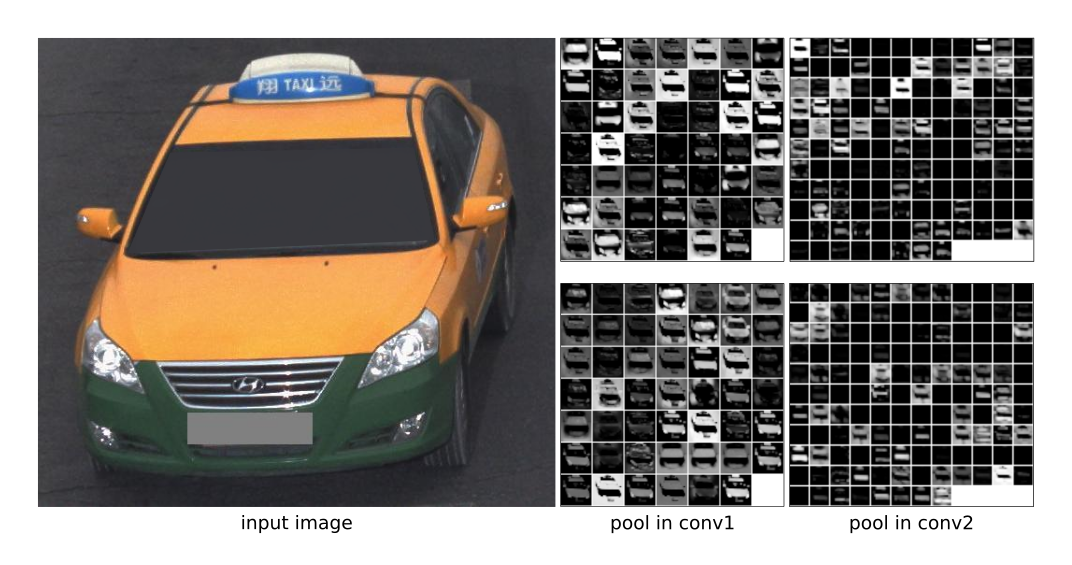}
		\label{fig:pooloutput}}
	}
	\caption{96 kernel with 11x11@3 resolution learned by our first convolutional layer with input resolution 224x224@3 and output example from pooling process. (a) 48 kernel from network 1, (b) 48 kernel from network 2, (c) output from pooling process in layer conv1 and conv2.}
	\label{fig:first_conv}
\end{figure*}

Another issue to tackle is execution time used to classify vehicle color. We implement the models using two different hardware, the first one the model running on 1 core CPU and the second one the model running on 448 cores GPU with NVIDIA Tesla C2050. Table \ref{tbl:tbl_exec} summarize average execution time for all testing examples. As shown in table \ref{tbl:tbl_exec}, the models that run on GPU have more than 20x faster than the models that run on CPU, so the issue of execution time is solved if the models running on appropriate hardware configuration. The initialization time is a time for the system to prepare the model, load it to the memory, and load mean image. For the practical implementation, we recommend using the client server mechanism, send the vehicle detection result to the server, do the vehicle color classification in server backend using GPU hardware, and send back the result to the Intelligent Transportation System for further processing.

To see how our models capturing color information in the data, we visualize a several layer of our CNN models. The first convolutional layer is an important part of the network to extract low-level features. Figure \ref{fig:first_conv} is a visualization of all kernels in the first convolutional layer and an example output of the pooling process in layer conv1 and conv2 of our CNN architecture. As seen in figure \ref{fig:first_conv}, the first convolutional layer capture rich color features in the input image. 
All vehicle color variations in dataset are present in the kernels. The kernels from network 1, figure \ref{fig:n1_conv1}, capture a lot of cyan-like color. Cyan-like color that appears in the kernel may contribute to the red color class or cyan color class. Another color that appears repeatedly in the kernel are red-blue color, green-gray color, and orange-like color. 
For further investigation, we capture respond from convolutional layer continuing with normalization and pooling process and it can see in figure \ref{fig:pooloutput}. We test our models using one of the test images and try to analyze the behaviour of our models. Figure \ref{fig:pooloutput} show that for yellow color class a lot of the green-like color kernel neuron is active and its looks like our models learned that color can be recognize from the hood color or the top color of the car. This behaviour occurs because most all of the images in dataset take image of the front of the car from some height and a little deviation of angle, so the side of the car is not cover very much. The camera configuration of taken images in dataset simulate the CCTV or other street camera that relatively used such that configuration. 

\section{Conclusion}
In the paper, we present the vehicle color recognition system using convolutional neural network. Our model succesfully capturing vehicle color in very high accuracy, 94,47\%, and outperform the original system provide by Chen \cite{pchen}. From the experiment, the best accuracy is achieve using RGB color space and this is contradictive with several papers that not recomend RGB color space for color recognition and using another color space like HSV or YUV. Execution time for our models is about 3 s for CPU (1 core) and 0.156 s for GPU (448 cores), although the execution time is slower than system provide by Chen \cite{pchen} but its still can be used for practical implementation with several adjustment.

%



%

\end{document}